\documentclass[conference]{IEEEtran}
\IEEEoverridecommandlockouts

\usepackage{cite}
\usepackage{amsmath,amssymb,amsfonts}
\usepackage{algorithmic}
\usepackage{graphicx}
\usepackage{textcomp}
\usepackage{xcolor}
\usepackage{adjustbox}
\usepackage{multicol}
\usepackage{multirow}
\usepackage{makecell}
\usepackage{hyperref}
\def\BibTeX{{\rm B\kern-.05em{\sc i\kern-.025em b}\kern-.08em
    T\kern-.1667em\lower.7ex\hbox{E}\kern-.125emX}}
\begin{document}

\title{\fontsize{20}{22}\selectfont MSA-ASR: Efficient Multilingual Speaker Attribution \\ with frozen ASR Models}

\author{\IEEEauthorblockN{Thai-Binh Nguyen}
\IEEEauthorblockA{
\textit{Karlsruhe Institute of Technology}\\
Karlsruhe, Germany \\
thai-binh.nguyen@kit.edu}
\and
\IEEEauthorblockN{Alexander Waibel}
\IEEEauthorblockA{
\textit{Carnegie Mellon University}\\
Pennsylvania, USA \\
alexander.waibel@cmu.edu}
}

\maketitle

\begin{abstract}
Speaker-attributed automatic speech recognition (SA-ASR) aims to transcribe speech while assigning transcripts to the corresponding speakers accurately. Existing methods often rely on complex modular systems or require extensive fine-tuning of joint modules, limiting their adaptability and general efficiency. This paper introduces a novel approach, leveraging a frozen multilingual ASR model to incorporate speaker attribution into the transcriptions, using only standard monolingual ASR datasets. Our method involves training a speaker module to predict speaker embeddings based on weak labels without requiring additional ASR model modifications. Despite being trained exclusively with non-overlapping monolingual data, our approach effectively extracts speaker attributes across diverse multilingual datasets, including those with overlapping speech. Experimental results demonstrate competitive performance compared to strong baselines, highlighting the model's robustness and potential for practical applications.
\end{abstract}

\begin{IEEEkeywords}
speaker-attributed, asr, multilingual
\end{IEEEkeywords}

\section{Introduction}
\label{section:introduction}

Speaker-attributed automatic speech recognition (SA-ASR) involves transcribing all speech within a multi-speaker recording and accurately attributing each spoken word to the correct speaker. Specifically, suppose a recording $X$ contains speech from $K$ different speakers. In that case, the objective is to generate a set of transcriptions $Y = \{y_1, y_2, ..., y_K\}$, where each $y_k$ corresponds to the sequence of words spoken by the speaker $k$. Addressing this challenge is crucial for various applications, from meeting transcription \cite{yang1999multimodal, gross2000towards, stiefelhagen2001estimating, 10.1145/971478.971505} to conversational AI \cite{waibel2010computers, waibel2012simultaneous}, where accurate speaker attribution is essential. Current approaches to this problem typically fall into two categories: modular strategies, which break down the task into separate components, and joint methods, which attempt to solve the problem in a unified framework.

Modular SA-ASR systems \cite{9383556, 9687974, Cui2024ImprovingSA, yu22b_interspeech, 10626098} decompose the SA-ASR task into sequential processing stages, typically involving speech separation, speaker diarization, and target speaker voice activity detection. These systems often assign speaker labels to speech segments prior to ASR. While modularity offers flexibility and the potential to leverage advancements in individual components, it can suffer from suboptimal performance due to the independent training of modules. Misalignments between the training objectives of these components can hinder the overall system's efficacy. In contrast, joint SA-ASR systems \cite{9687974, 9383600} address these limitations by processing the entire SA-ASR task end-to-end, potentially achieving improved performance and coherence.

Joint SA-ASR systems, also known as E2E SA-ASR models, typically consist of two main components: the ASR module and the Speaker module. The ASR module generally generates a sequence of tokens, while the Speaker module produces speaker embeddings for each token. These embeddings are then used for speaker identification \cite{9746225, Kanda2022StreamingSA}. In studies such as \cite{Li2023SaParaformerNE, kanda21b_interspeech, shi23d_interspeech}, speaker embeddings have been shown to enhance ASR performance by providing additional features for the ASR decoder layer. In some cases \cite{Cornell2023OneMT}, rather than outputting speaker embeddings explicitly, they are utilized to help the ASR directly generate speaker labels. Often \cite{Kanda2022StreamingSA, 9383600, Cornell2023OneMT, Li2023SaParaformerNE}, the ASR module not only generates a sequence of tokens but also produces special tokens (e.g., $<cc>, <sc>$) to indicate a change in speaker.

While much research on E2E SA-ASR has centered on enhancing ASR performance by incorporating speaker information, our approach offers a different perspective. Traditional joint models often involve fine-tuning the ASR component to add capabilities like generating speaker change tokens or managing overlapping speech, usually relying on limited, language-specific datasets. However, since overlapping speech constitutes roughly 10\% of multi-talker data (as illustrated in Table \ref{tab:data-overlapping}) and given the success of models like Whisper, which are trained on large and diverse datasets, we question the necessity of extensive ASR fine-tuning. The Whisper study \cite{radford2022robust} demonstrates that models trained on comprehensive and varied datasets can effectively handle a wide range of speech recognition tasks. In contrast, models tailored to specific datasets may excel within those domains but often lack broader robustness. We suggest that by focusing on the speaker module and utilizing a robust pre-trained ASR model, effective SA-ASR can be achieved without compromising generalizability.

To demonstrate generalizability, we extend SA-ASR to handle multilingual speech. Multilingual SA-ASR studies are rare, with the last in 2002 on speaker ID via multilingual phone strings \cite{jin2002speaker, jin2002phonetic}. Research in zero-shot multilingual transfer learning \cite{pires-etal-2019-multilingual} has shown that a frozen multilingual pre-trained model can be trained for a task in one language and then used to make predictions in another. Additionally, studies \cite{10446589, Yang2022SimulatingRS} have explored training models for multi-speaker speech recognition using standard ASR datasets. Building on this foundation, we leverage the Whisper model's capacity to process multiple languages and follow the approach suggested by \cite{10446589} to adapt regular English ASR datasets. This allows us to create a new E2E SA-ASR model (MSA-ASR) that can predict speakers across different languages, introducing a novel method for multilingual SA-ASR. Through benchmarking across various datasets, we demonstrate the effectiveness of our proposed approach in handling SA-ASR tasks across different languages and conditions.

\begin{table}[]
\centering
\begin{tabular}{lrr}
\hline
Dataset          & Overlap (\%) & Duration (hours) \\ \hline
ESTER 1\&2 \cite{lebourdais-etal-2022-overlaps}        & 0.67         & 260              \\
ETAPE \cite{lebourdais-etal-2022-overlaps}             & 5.29         & 105              \\
EPAC \cite{lebourdais-etal-2022-overlaps}              & 1.11         & 34               \\
REPERE \cite{lebourdais-etal-2022-overlaps}            & 3.36         & 58               \\
DIHARD \cite{lebourdais-etal-2022-overlaps}            & 11.6         & 34               \\
AMI \cite{lebourdais-etal-2022-overlaps}          & 13.87        & 96               \\
FISHER \cite{Abdullah_2017}          & 13.53        & 984              \\
CHIME 6 dev/eval \cite{cornell20_interspeech} & 21 / 15      & 4.5 / 5.1        \\ \hline
\end{tabular}
\vspace{3mm}
\caption{Total duration and proportion of overlaps duration for different speech corpora.}
\label{tab:data-overlapping}
\vspace{-2.5em}
\end{table}

\section{Approach}
\label{section:approach}
\subsection{Modeling}
\label{sub:modeling}

\begin{figure}[htpb]
  \centering
  \includegraphics[width=1.0\linewidth]{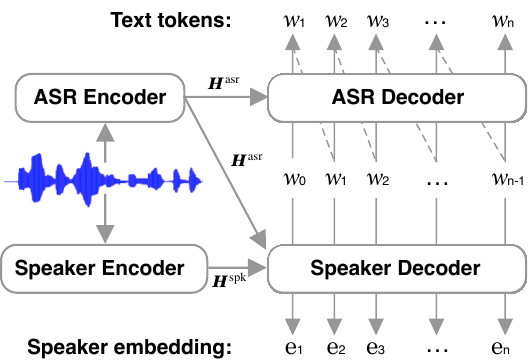}

  \vspace{-0.5em}
    \caption{Overview of MSA-ASR model. ASR decoder processes tokens sequentially during inference, while the Speaker decoder can generate speaker embeddings in parallel.}
  \label{fig:model_overview}
\vspace{-0.5em}
\end{figure}

In designing our MSA-ASR model (Figure \ref{fig:model_overview}), we sought to blend the advantages of both modular and joint SA-ASR systems. The model consists of two main components: the ASR and Speaker modules. By keeping the frozen ASR module as a modular system, we ensure the speech recognition process remains stable and generalizable across diverse languages and domains. Meanwhile, the Speaker module is fine-tuned to work in harmony with the ASR system as a joint system.

ASR module is a transformer seq2seq model containing an encoder and decoder. The encoder takes the input signal $X$ to produce hidden features $H^{\rm asr}\in\mathbb{R}^{L\times f^e}$ where $f^e$ and $L$ are the feature dimension and the length of the feature sequence. The ASR decoder then iterative estimates sequence $W = [w_1,..., w_N]$. At each decoder step, the ASR decoder calculates the output $w_n = ASRDecoder(w_{[0:n-1]}, H^{\rm asr})\in\mathcal{V}$ ($\mathcal{V}$ is the ASR vocabulary) given previous token $w_{[0:n-1]}$ and encoder hidden features $ H^{\rm asr}$.

\begin{figure}[htpb]
  \centering
  \includegraphics[width=1.0\linewidth]{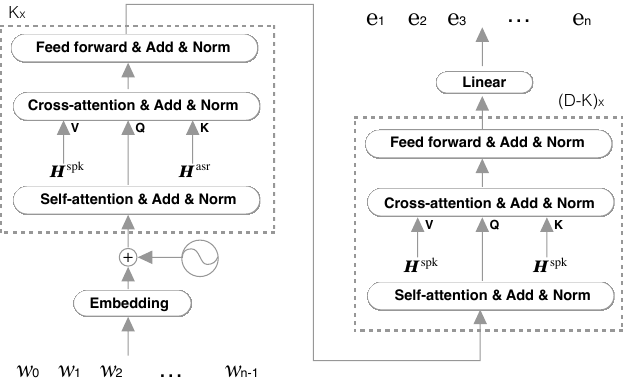}

  \vspace{-0.5em}
    \caption{Speaker decoder architecture. Similar as standard transformer decoder, but cross-attention uses key and value from a different encoders.}
  \label{fig:model_spk_decoder}
\vspace{-1.5em}
\end{figure}

The Speaker module predicts a sequence of speaker embeddings $E = [e_1, e_2, ..., e_N]\in\mathbb{R}^{N\times f^d}$ where $f^d$ denotes the speaker embedding dimension. Each embedding $e_n$ corresponds to a token $w_n$ in the ASR-generated sequence. This module, like the ASR, uses a transformer architecture. Speaker encoder transforms $X$ into hidden features $H^{\rm spk}\in\mathbb{R}^{L\times f^e}$ (same shape as ASR encoder output). Figure \ref{fig:model_spk_decoder} illustrates the architecture of the Speaker decoder. The word and position embeddings are shared between the ASR and Speaker decoder modules to ensure alignment between their outputs. The cross-attention of the first $K$ layers in the Speaker decoder has been customized to take $H^{\rm asr}$ as the key, $H^{\rm spk}$ as the value. In the rest of $(D-K)$ layers, the key has been calculated from $H^{\rm spk}$. 

The Speaker module's training objective is to optimize $E$ to closely match the target speaker embedding sequence $T = [t_1, t_2, ..., t_N]\in\mathbb{R}^{N\times f^d}$ using cosine similarity loss \eqref{eq:loss_1}. We also want the model to distinguish between speakers inside an utterance. To do this, we first calculate the pairwise cosine similarity between the embeddings within the output sequence $E$ (denoted as $C_{ee}$) \eqref{eq:loss_ooc}, between the embeddings within the target sequence $T$ (denoted as $C_{tt}$) \eqref{eq:loss_ttc}, also the pairwise cosine similarity between $E$ and $T$ (denoted as $C_{et}$) \eqref{eq:loss_otc}. Then an MSE loss (\ref{eq:loss_2}, \ref{eq:loss_3}) used to make both $C_{ee}$ and $C_{et}$ close to $C_{tt}$. Our final Embedding Alignment and Discrimination loss (EAD) will be the weighted sum of 3 losses \eqref{eq:loss_final}.

\vspace{-3mm}
\begin{align}
C_{ee} &= [cos(e_i, e_j)]_{i,j=1}^N \in \mathbb{R}^{N \times N}  \hspace{0.2cm} \text{\scriptsize{pairwise similarity within \textit{E}}} \label{eq:loss_ooc} \\
C_{et} &= [cos(e_i, t_j)]_{i,j=1}^N \in \mathbb{R}^{N \times N} \hspace{0.2cm} \text{\scriptsize{between \textit{E} and \textit{T}}} \label{eq:loss_otc} \\
C_{tt} &= [cos(t_i, t_j)]_{i,j=1}^N \in \mathbb{R}^{N \times N} \hspace{0.2cm} \text{\scriptsize{within \textit{T}}} \label{eq:loss_ttc} \\
L_1 &= \sum_{i=1}^{N} (1 - cos(t_i, e_i)) \in \mathbb{R} \hspace{0.2cm} \text{\scriptsize{cosine similarity loss with E, T}} \label{eq:loss_1} \\
L_2 &= \frac{1}{N^2} \sum_{i=1}^{N} \sum_{j=1}^{N} (C_{ee}[i,j] - C_{tt}[i,j])^2 \in \mathbb{R} \hspace{0.3cm} \text{\scriptsize{MSE loss}} \label{eq:loss_2} \\
L_3 &= \frac{1}{N^2} \sum_{i=1}^{N} \sum_{j=1}^{N} (C_{et}[i,j] - C_{tt}[i,j])^2 \in \mathbb{R} \hspace{0.3cm} \text{\scriptsize{MSE loss}} \label{eq:loss_3} \\
L &= \alpha L_1 + \beta L_2 + \gamma L_3 \hspace{0.5cm} \text{\scriptsize{EAD Loss}}\label{eq:loss_final}
\end{align}

Our EAD loss function is similar to triplet and contrastive loss (widely used for speaker verification tasks \cite{coria2020comparison}), as all aim to distinguish between different speakers in the embedding space. Like triplet loss, which pulls similar samples together and pushes dissimilar ones apart, and contrastive loss, which minimizes distances within similar pairs, our loss uses cosine similarity ($L_1$) to align output and target embeddings while maintaining internal pairwise relationships ($L_2, L_3$). Our EAD loss is chosen because we don't have the ground truth of speaker labels but only the target speaker embedding $T$, so these target functions help in effective embedding differentiation and alignment without needing explicit labels. A detail of how $T$ has been constructed is presented in the next section.

\subsection{Data Processing}
\label{sub:data_processing}

We extend the approach outlined in \cite{10446589}, which processed speaker turns in conventional ASR datasets for multi-talker speech recognition. While the previous work focused on detecting speaker changes by combining random turns, it did not account for the need to extract consistent speaker embeddings, as each speaker might only appear once, limiting the model's ability to differentiate between speakers.

To overcome this, we employed a pre-trained speaker embedding model, TitaNet-L \cite{9746806}, trained on 7,000 hours of diverse speech datasets, to compute an embedding for each speaker's turn. Since clustering turns from the same speaker is inherently complex, we opted not to rely on cluster labels. Instead, as detailed in section \ref{sub:modeling}, we used the original speaker embeddings as weak labels, avoiding the need for explicit speaker labels. We identified similar turns by selecting those with a cosine similarity score above the threshold $\theta$.

The training samples were created by pairing turns randomly. The target sequence of speaker embeddings $T$ is constructed by aligning each turn's transcript with its corresponding speaker embedding. Each turn was paired with at least one similar counterpart, and we limited each sample to five distinct turn groups with no overlapping similar turns. This process ensured that each sample contained up to 30 seconds of no overlapping speech from a maximum of five speakers. Similar to \cite{10446589}, we also incorporated random noise and reverberation to enhance the data.

\section{Experiments}
\subsection{Datasets}
\label{sub:datasets}

To benchmark systems, we utilize three types of datasets: multilingual (Voxpopuli \cite{wang-etal-2021-voxpopuli}), monolingual (AMI-IHM \cite{carletta2005ami}, LibriCSS \cite{9053426}), and mixed-language (in-house data). Since AMI-IHM and LibriCSS are typically employed for benchmarking English SA-ASR systems, we will not delve into their details here. The processing for the other datasets is as follows:

Voxpopuli is a multilingual speech corpus featuring one speaker per sample across 16 languages. To adapt it for the multi-talker benchmark, we used the approach from \cite{10446589}, where test set utterances are randomly concatenated. This produces samples with an average of 2.5 speakers, 20 seconds of audio, and up to 5 non-overlapping turns.

In our study, a mixed-language dataset includes samples with multiple languages. Our dataset features English, German, Turkish, and Vietnamese. Each session involves two speakers discussing a scientific paper, one speaking in English and the other in one of the other languages. The dataset totals 45 minutes, with language distribution as follows: English (44\%), German (9\%), Turkish (12\%), and Vietnamese (35\%). The overlap rate is approximately 3\%.

\begin{table}[]
\centering
\fontsize{9pt}{9pt}\selectfont
\begin{tabular}{lccc}
\hline
\textbf{Language} & \textbf{Diarization+ASR} & \textbf{MSA-ASR (Our)} & \textbf{ASR} \\ \hline
English           & 15.52                    & 12.90                 & 12.24        \\
German            & 26.24                    & 16.54                 & 14.28        \\
French            & 32.20                    & 16.53                 & 13.95        \\
Spanish           & 20.75                    & 13.73                 & 11.32        \\
Polish            & 34.94                    & 16.15                 & 10.31        \\
Italian           & 33.52                    & 23.71                 & 20.06        \\
Romanian          & 38.48                    & 23.65                 & 18.05        \\
Hungarian         & 29.77                    & 28.12                 & 19.82        \\
Czech             & 35.47                    & 28.60                 & 16.27        \\
Dutch             & 29.81                    & 18.12                 & 14.68        \\
Finnish           & 37.30                    & 20.51                 & 15.85        \\
Croatian          & 37.52                    & 34.52                 & 28.05        \\
Slovak            & 34.44                    & 27.24                 & 16.07        \\
Slovenian         & 41.32                    & 30.90                 & 27.33        \\
Estonian          & 44.65                    & 39.59                 & 37.15        \\
Lithuanian        & 69.17                    & 40.57                 & 34.04        \\ \hline
\end{tabular}
\vspace{3mm}
\caption{Comparison of cpWER (\%) across multiple languages for non-overlapping multi-talker Voxpopuli using diarization with ASR (Pyannote + Whisper large-v2), our SA-ASR system, and ASR without speaker consideration.}
\label{tab:cpwer_voxpopuli}
\vspace{-2.5em}
\end{table}

\subsection{Modeling and Metric}

All systems will be evaluated using the concatenated minimum permutation word error rate (cpWER) \cite{watanabe2020chime}, depending on ASR performance and speaker labels. For the multilingual and mixed-language datasets, the baseline system is Diarization + ASR where diarization is Pyannote 3.1 \cite{Bredin23} and ASR is Whisper large-v2. The baseline for the monolingual English dataset will vary between modular and joint systems, as outlined in section \ref{section:introduction}.

Our MSA-ASR model employs Whisper large-v2 as the ASR component. The Speaker model has 12 layers for each encoder and decoder. The first $K=1$ layers of the Speaker decoder use $H^{\rm asr}$ as the key. $\alpha=\beta=\gamma=1$ for the EAD loss. $\theta = 0.7$ is used for grouping similar speaker turns. We train this model for 250,000 steps with batch size 80 (equivalent to 40 minutes of audio), using AdamW optimization with a learning rate 1e-4. We utilize spectral clustering \cite{wang2018speaker} for speaker assignment. Our MSA-ASR model is only trained on the data that has been processed as described in section \ref{sub:data_processing} without fine-tuning for the in-domain data (section \ref{sub:datasets}).

\begin{table*}[t]
    \begin{minipage}{.56\linewidth}
\caption{Comparison of cpWER (\%) for LibriCSS}
\label{tab:cpwer_libricss}
      \vspace{-0.5em}
      \centering
      \adjustbox{max width=\linewidth}{%

\begin{tabular}{l|cccccc}
\hline
\multicolumn{1}{c|}{\multirow{2}{*}{System}} & \multicolumn{6}{c}{Overlap ratio in \%}                                                    \\
\multicolumn{1}{c|}{}                        & 0S           & 0L           & 10           & 20            & 30            & 40            \\  
 \hline 
LSTM SOT-SA-ASR \cite{9383600}                                  & 10.3         & 15.8         & 13.4         & 17.1          & 24.4          & 28.6          \\
Conformer SOT-SA-ASR\cite{kanda21b_interspeech}                            & 8.6          & 12.7         & 11.2         & 11.3          & 16.1          & 17.5          \\
TS-VAD + ASR \cite{9383556}                                & 9.5          & 11.0         & 16.1         & 23.1          & 33.8          & 40.9          \\
NME-SC + SOT-SA-ASR \cite{9746225}                             & 9.0          & 12.2         & \textbf{8.7} & \textbf{10.9} & \textbf{13.7} & \textbf{13.9} \\
MSA-ASR (our)                                 & \textbf{7.5} & \textbf{8.1} & 11.5         & 27.9          & 41.7          & 46.5          \\ \hline
\end{tabular}}
    
    \end{minipage}%
    \hfill
    \begin{minipage}{.42\linewidth}
      \centering
\caption{Comparison of cpWER (\%) for AMI-IHM. All systems use gold VAD.}
\label{tab:cpwer_amiihm}
        \vspace{1em}
        \adjustbox{max width=\linewidth}{%
\begin{tabular}{l|cc}
\hline
System                         & Dev  & Eval \\ \hline
Transformer SOT-SA-ASR  \cite{9687974}                       & \textbf{14.5} & 15.0 \\
NME-SC + SOT-SA-ASR  \cite{9746225}              & 16.3 & 15.1 \\
MSA-ASR (our)                   & 15.6 & \textbf{14.3} \\
Gold transcript + MSA-ASR (our) & 2.7  & 1.9  \\ \hline
\end{tabular}}
    \end{minipage}%

    \vspace{-1em}
\end{table*}

\subsection{Results}

Table \ref{tab:cpwer_voxpopuli} shows the benchmark result on the multi-talker Voxpopuli dataset. The first column is the results of using diarization followed by ASR. The second column shows the performance of our system. The third column is the baseline ASR performance, which does not account for speaker labels. This baseline ASR value is the lower bound WER, as the other systems incorporate speaker information into the ASR output. Overall, our system introduces a 29.3\% relative error increase over the baseline ASR, whereas the Diarization + ASR system increases it by 92\%. 

Although both systems perform well in English, for languages with large datasets (over 1,000 hours, detail in whisper paper \cite{radford2022robust}) used to fine-tune the ASR model, such as German, French, Spanish, Polish, Italian, Dutch, and Finnish, our system results in a 26\% relative error increase, compared to a 120\% increase with Diarization + ASR. For languages with smaller datasets, including Romanian, Hungarian, Czech, Croatian, Slovak, Slovenian, Estonian, and Lithuanian, our system introduces a 35\% relative error increase, while Diarization + ASR results in a 76\% increase. These results demonstrate that, compared to the language-independent diarization approach, using the same ASR model, our system more effectively leverages the generalizability of the ASR model to handle multilingual scenarios, particularly in languages with larger datasets.

Table \ref{tab:cpwer_libricss} compares different systems on the LibriCSS dataset across various overlapping ratios. All systems, except TS-VAD, utilize a VAD model to segment long audio into smaller chunks. In our setting, we use Silero VAD\cite{Silero_VAD}. The LSTM SOT SA-ASR \cite{9383600} and Conformer SOT SA-ASR \cite{kanda21b_interspeech} systems are similar to ours but enhance the ASR model by incorporating speaker embeddings. TS-VAD is a target speaker voice activity detection system, which, in \cite{9383556}, is followed by an ASR model to transcribe the detected speaker. NME-SC is a diarization system that integrates Conformer SOT SA-ASR as in \cite{9746225}. Our MSA-ASR system outperforms the others in scenarios without overlapping speech but shows decreased performance as the overlap ratio increases. This is expected, as our ASR model is frozen, whereas other systems are fine-tuned for this specific data type.

One of the significant advantages of our MSA-ASR model is its ability to run the Speaker model independently of the ASR model. This feature provides greater flexibility and efficiency in processing. Table \ref{tab:cpwer_amiihm} presents benchmark results for various joint systems on the AMI-IHM dataset, using the ideal scenario where gold VAD labels are available. In this scenario, all systems focus solely on ASR and assigning speaker information. Our MSA-ASR model, even without fine-tuning on AMI-IHM, demonstrates competitive performance compared to other state-of-the-art models like the Transformer SOT SA-ASR \cite{9687974} and NME-SC + SOT SA-ASR \cite{9746225}. The final row of table \ref{tab:cpwer_amiihm} highlights our model's unique capability to directly accept gold transcripts for assigning speaker embeddings, resulting in exceptional performance.

Table \ref{tab:cpwer_kit} highlights the performance of our system on real multilingual meeting data from a mixed-language dataset. The long audio recordings were segmented into smaller chunks using Silero VAD and then processed using our MSA-ASR model. Since Whisper large-v2 is a multilingual ASR model, it effectively manages multilingual audio. Compared to the Diarization + ASR system, our MSA-ASR model delivers significantly better results. Error analysis shows that while both ASR and diarization have acceptable error rates, averaging 7.67\% and 7.38\%, respectively, combining diarization with ASR leads to a higher cpWER than using a joint ASR and speaker model like our MSA-ASR. 

Table \ref{tab:cpwer_voxpopuli} and table \ref{tab:cpwer_kit} highlight the advantages of our joint speaker attribute system, which effectively handles multilingual scenarios despite being fine-tuned only on an English dataset. While other joint systems often need fine-tuning for each specific language (due to data constraints) and are limited to those trained languages, our system demonstrates the ability to generalize and adapt to multiple languages without additional fine-tuning.

\begin{table}[]
\centering
\begin{tabular}{lcccc}
\hline
System          & en\_de & en\_tr & en\_vi & avg \\ \hline
Diarization + ASR & 6.81   & 15.76  & 18.54  & 13.70       \\
MSA-ASR (our)    & 5.71   & 5.98   & 11.54  & 7.74       \\ \hline
\end{tabular}
\vspace{1em}
\caption{Comparison of cpWER (\%) for mix-languages meeting dataset.}
\label{tab:cpwer_kit}
\vspace{-3.5em}
\end{table}

\section{Conclusion}
\vspace{-0.3em}

This study introduces a novel approach for integrating ASR and Speaker models into a unified speaker-attribute speech recognition system, capable of handling multilingual datasets while using only standard monolingual ASR data. By fine-tuning exclusively on the speaker model, we preserve the original performance of the ASR model. Our system, although primarily optimized for non-overlapping data, also demonstrates robust performance across a range of diverse benchmarks. This highlights its capability to manage real-world scenarios with varying complexities effectively. We have made our pre-trained model and dataset publicly available for further research at \url{hf.co/nguyenvulebinh/MSA-ASR}.

\section{Acknowledgment}
\vspace{-0.2em}

The authors gratefully acknowledge support from Carl Zeiss Stiftung under the project Jung bleiben mit Robotern (P2019-01-002). This work was also partially supported by the European Union’s Horizon research and innovation programme (grant No. 101135798, project Meetween), the Volkswagen Foundation project ``How is AI Changing Science? Research in the Era of Learning Algorithms'' (HiAICS), and KIT Campus Transfer GmbH (KCT) staff in accordance to the collaboration with Carnegie-AI.

\bibliographystyle{IEEEtran}
\bibliography{IEEEabrv,mybib}

\end{document}